\documentclass[10pt,twocolumn,letterpaper]{article}

\usepackage[pagenumbers]{cvpr} 

\usepackage[dvipsnames]{xcolor}

\definecolor{cvprblue}{rgb}{0.21,0.49,0.74}
\usepackage[pagebackref,breaklinks,colorlinks,citecolor=cvprblue]{hyperref}

\usepackage{amsmath}
\usepackage{breqn}
\usepackage{algorithm}
\usepackage{algpseudocode}
\usepackage{booktabs}
\usepackage{float}

\usepackage{amssymb}
\usepackage{hyperref}

\usepackage{breqn}
\usepackage{placeins}
\usepackage{tabularx}
\usepackage{multirow}
\usepackage{colortbl}
\usepackage{tabularx}
\usepackage{caption}
\usepackage{tablefootnote}

\def\paperID{8724} 
\def\confName{CVPR}
\def\confYear{2024}

\title{Diffusion-based Data Augmentation for Object Counting Problems}

\author{
Zhen Wang$^1$$^*$
\and
Yuelei Li$^1$$^*$
\and
Jia Wan$^2$
\and 
Nuno Vasconcelos$^1$\\
\and
$^1$University of California, San Diego
\and
$^2$Harbin Institute of Technology (Shenzhen)
}

\begin{document}

\twocolumn[{%
\renewcommand\twocolumn[1][]{#1}%
\maketitle
\begin{center}
    \centering
    \captionsetup{type=figure}
    \includegraphics[width=0.95\textwidth]{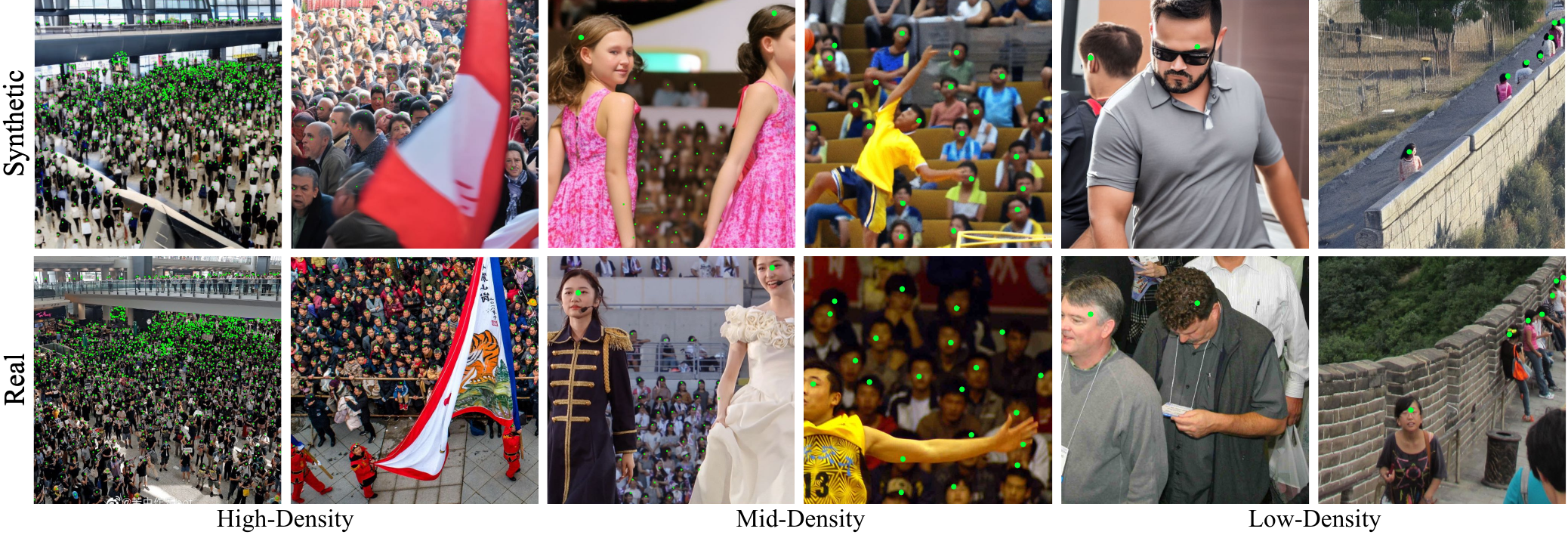}
    \captionof{figure}{\textbf{Generated images in various crowd density levels}: Our model is able to generate crowd images from head location maps with various density levels, as illustrated in this figure. {Please zoom in for details.}}
    \label{fig:density}
\end{center}%
}]
\def\thefootnote{*}\footnotetext{These authors contributed equally to this work}\def\thefootnote{\arabic{footnote}}

\begin{abstract}
Crowd counting is an important problem in computer vision due to its wide range of applications in image understanding. Currently, this problem is typically addressed using deep learning approaches, such as Convolutional Neural Networks (CNNs) and Transformers. However, deep networks are data-driven and are prone to overfitting, especially when the available labeled crowd dataset is limited. To overcome this limitation, we have designed a pipeline that utilizes a diffusion model to generate extensive training data. We are the first to generate images conditioned on a location dot map (a binary dot map that specifies the location of human heads) with a diffusion model. We are also the first to use these diverse synthetic data to augment the crowd counting models. Our proposed smoothed density map input for ControlNet significantly improves ControlNet's performance in generating crowds in the correct locations. Also, Our proposed counting loss for the diffusion model effectively minimizes the discrepancies between the location dot map and the crowd images generated. Additionally, our innovative guidance sampling further directs the diffusion process toward regions where the generated crowd images align most accurately with the location dot map. Collectively, we have enhanced ControlNet's ability to generate specified objects from a location dot map, which can be used for data augmentation in various counting problems. Moreover, our framework is versatile and can be easily adapted to all kinds of counting problems. Extensive experiments demonstrate that our framework improves the counting performance on the ShanghaiTech, NWPU-Crowd, UCF-QNRF, and TRANCOS datasets, showcasing its effectiveness.
\end{abstract}    
\section{Introduction}
\label{sec:intro}

Dense crowd counting is an important technology for public safety and surveillance \cite{crowdanalysis}. In recent years, various learning-based methods \cite{Chen_2023,ranasinghe2023diffusedenoisecount, huang2023counting, liang2022endtoend, han2023steerer} have been proposed to address this problem. With the increase in the popularity of deep learning models, this has created the need for large datasets.  
However, because it is extremely labor intensive to label dense crowd images manually, publicly available labeled crowd counting datasets \cite{gao2020nwpu, sindagi2020jhu-crowd++, liu2018ano_pred} tend to be relatively small, with less than 3,500 images each. This induces model overfitting to the training data and can produce models that generalize poorly to unseen data \cite{overfitting}. 

Recently, generative models such as denoising diffusion probabilistic models (DDPM) and score-based generative models have been shown quite effective at image generation. Many works have introduced variants to control the generation process, by conditioning the generation on images~\cite{saharia2022palette,rombach2022highresolution}, text~\cite{saharia2022photorealistic}, sound~\cite{yariv2023audiotoken}, canny edges~\cite{zhang2023adding}, etc. In principle, such models could be used to synthesize crowd images that could be then used to augment crowd counting datasets. However, the synthesis of images of crowds requires conditioning based on the dot annotations commonly used in counting datasets, which has so far not been explored. 

In this work,  we propose a task-specific regularization to address this problem. We propose a method that augments the ControlNet \cite{zhang2023adding} to synthesize images conditioned on a location dot map, a binary map that specifies the locations of human heads in an image of a crowded scene. We convolve the location dot map with a small Gaussian kernel and use the resulting density map as input to ControlNet, which significantly improves ControlNet's performance in generating crowds at the correct locations. Also, the model is trained with a counting loss that further encourages the synthesized crowd to comply with the conditioning location dot map. This is complemented by a new form of counting guidance during the sampling process, which encourages the model to produce the most accurate images. These images are finally used to augment the training set of crowd counting models and improve performance on popular datasets.

While conditioned image generation has been applied to various tasks \cite{shi2023dragdiffusion, yang2022paint, kim2022diffusionclip}, to the best of our knowledge this is the first approach that is conditioned on position dot map. This is also the first attempt to integrate conditional generation as a data augmentation tool for the dense object counting problem. 
Beyond this, our contributions are: 
1) A highly adaptable data augmentation framework is proposed that can generate images with various scenes and crowd densities by modifying the conditioned text and location prompts.
2) A smoothed density map conditioning method is introduced to enhance ControlNet's performance in generating objects based on specific location dot maps, thereby improving the precision and relevance of the generated images.
4) An innovative counting loss and counting guided sampling method are employed in the diffusion model to further ensure the correspondence between the location maps and the resulting crowd images.
5) The proposed method has been applied to various objects like crowds and vehicles. The outcomes highlight the strong correspondence between the location maps and the synthesized images.

\section{Related Work}
\label{sec:related}

\subsection{Crowd Counting}
Crowd counting, a representative task in computer vision, confronts inherent difficulties due to scale variations, frequent occlusions, and the inherently dense nature of crowds. The pioneering multi-column neural network (MCNN) \cite{zhang2016single} was introduced to extract multi-scale features, while alternative solutions like the image pyramid \cite{kang2018crowd} have been employed to deal with scale variation challenges effectively. Enhancing generalization, the cross-scene crowd counting algorithm \cite{zhang2015cross} adapts models to new environments, and methods leveraging correlation information \cite{wan2019residual} have also shown promise. However, these methods typically rely on intermediate density maps. So, some novel strategies propose to use point annotations directly to refine accuracy \cite{DM-Count,Wan_2021_CVPR}. Furthermore, cutting-edge frameworks such as Transformers \cite{liang2022endtoend} and diffusion models \cite{ranasinghe2023diffusedenoisecount} are being investigated to further elevate counting accuracy.

All these complicated deep learning models are data-driven, which requires a large amount of labeled data during training. Therefore, the existing small counting datasets (less than 3500 training images each) are not enough for training large deep learning models and data augmentation is necessary in this situation.

\subsection{Data Augmentation}

The most commonly used methods in crowd counting problems are randomly flipping, rotating, cropping, resizing, etc \cite{yang2022image}. Those basic methods cannot provide enough variation to the training data to make the deep learning model more robust.
Works have been done on applying generative models, such as GAN and diffusion, to self-supervised learning and classification tasks. \cite{IDiff-Face_2023_ICCV} applied diffusion model for face recognition task that increases the intra-class data diversity during training;  \cite{diffusion_cell_aug} used diffusion model to augment the data for cell cycle phase classification. The medical AI community has also widely explored using GAN for medical data augmentation in classification tasks, e.g. \cite{gan_xray_aug, gan_skin_lesion_aug} using GAN for X-ray data and skin lesion data augmentation respectively.

Traditional data augmentation techniques fall short in modifying crowd distributions, as they're unable to create images with people at specific locations. The proposed pipeline, however, ensures that position annotations align with the generated images. It also leverages a large pre-trained diffusion model to produce high-quality crowd images, thus bypassing the need for intensive manual labeling.

\subsection{Diffusion Models}
Diffusion models are a family of probabilistic generative models that have achieved state-of-the-art results recently \cite{ho2020denoising}. Unlike previous models such as GANs \cite{goodfellow2014generative} that suffer from mode collapse and instability during training, the optimization of diffusion models is similar to a score matching and by \cite{ho2020denoising} the loss can simply be a Mean Squared Error (MSE).

However, using the diffusion model in our data augmentation task requires careful design because we do not have the ground truth crowd images, which means the original training objective of diffusion models cannot be directly used. Also, diffusion models require multiple steps in the denoising process to generate an image. Given the noise, it is hard to add the position constraints of the generated crowd in this process. In our work, we address these issues by introducing a new loss and a new sampling strategy. Leveraging the success of the large pretrained diffusion model, our pipeline can produce high-quality images with controllability. 
\begin{figure*}[t!]
    \centering
    \includegraphics[width=0.98\textwidth]{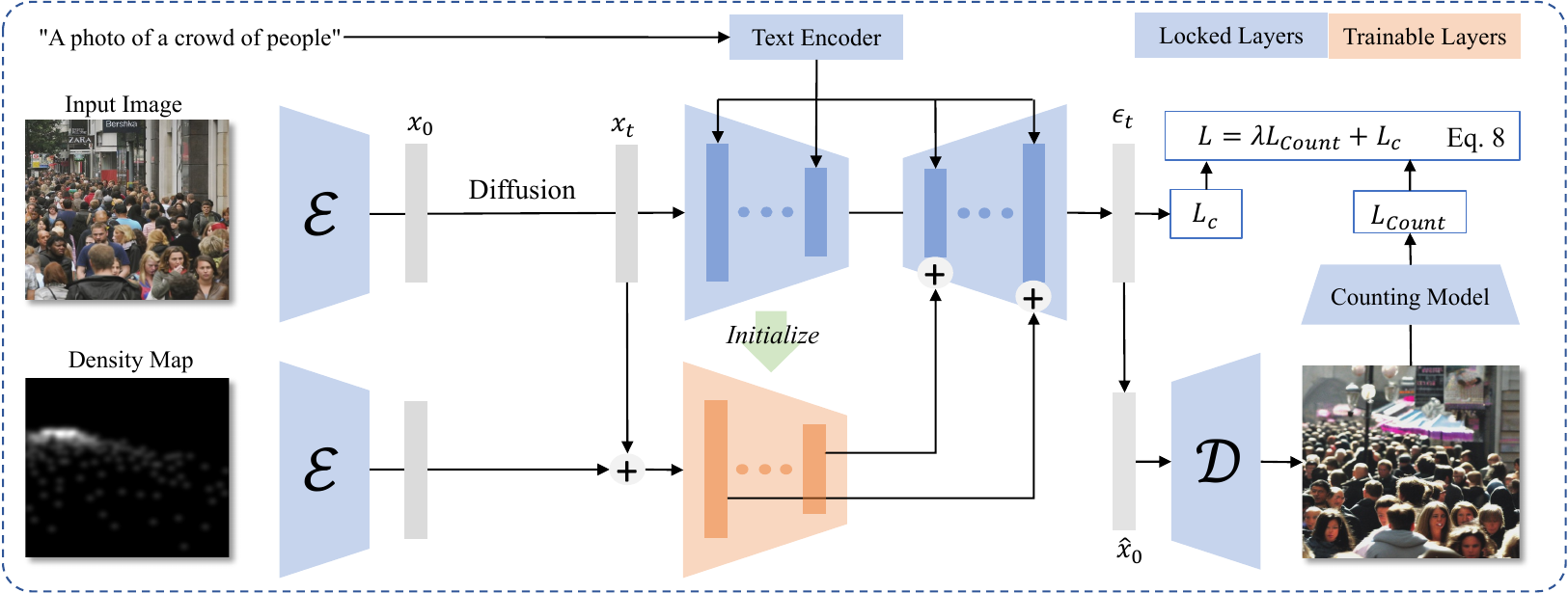}
    \caption{\textbf{Overview of Training Pipeline}: We propose a diffusion-based framework for data augmentation for the counting model. During training, we have $L_{\text{c}}$ loss to predict the gradient of log data density $\epsilon$. We also have a counting loss $L_{\text{Count}}$ to enforce the correspondence between the input dot map and the generated image.}
    \label{fig:training-pipeline}
    \vspace{-10pt}
\end{figure*}

\section{Method}
\label{sec:method}

In this section, we briefly review diffusion models and introduce the proposed labeling-free data augmentation framework.

\subsection{Background}\label{sec:diff}

\textbf{Diffusion Models.}
Diffusion models are a class of generative models that involve two processes: a forward diffusion process that adds noise to images, and a backward process that reconstructs these images from their noisy counterparts. The diffusion process is usually not implemented in the image space but in the latent space of an image auto-encoder. In what follows, we will use $x$ to denote either images or latent codes (if an autoencoder is used). The forward Markov process can be described as
\begin{equation}
    p(x_t|x_{t-1}) = \mathcal{N}(x_t ; \sqrt{1 - \beta_t}x_{t-1}, \beta_t I), 
\end{equation}
where $\beta_t$ is the variance of Gaussian noise at time step $t$. Based on this, the distribution of $x_t$ given $x_0$ is
\begin{equation}
    p(x_t|x_0) = \mathcal{N}(x_t; \sqrt{\Bar{\alpha}_t}x_0, (1-\Bar{\alpha}_t)I),
\end{equation}
where $\Bar{\alpha_t} = \prod_{i=1}^t \alpha_i$ and $\alpha_i = 1 - \beta_i.$ For large $t$, $\Bar{\alpha_t} \approx 0$ and this converges to a white Gaussian noise code $x_T$, where $T$ is the number of diffusion steps, typically $T=1000$ in our experiments. At time $t$, the noisy sample $x_t$ is related to the original image by
\begin{equation}
    x_t = \sqrt{\Bar{\alpha}_t}x_0 + \sqrt{(1 - \Bar{\alpha}_t)} \epsilon
    \label{eq:xt}
\end{equation}
where $\epsilon$ is a sample from an i.i.d. zero-mean Gaussian variable. 
In the backward process, a neural network $\epsilon_\theta$ is trained to denoise $x_T$, by learning to predict the noise $\epsilon$ added at each step of the forward process, which reduces to minimizing the loss
\begin{equation}
L_{u} = \|\epsilon - \epsilon_\theta(x_t,t)\|^2.
\label{eq:Lu}
\end{equation}

\begin{figure*}[t!]
    \centering
    \includegraphics[width=0.98\textwidth]{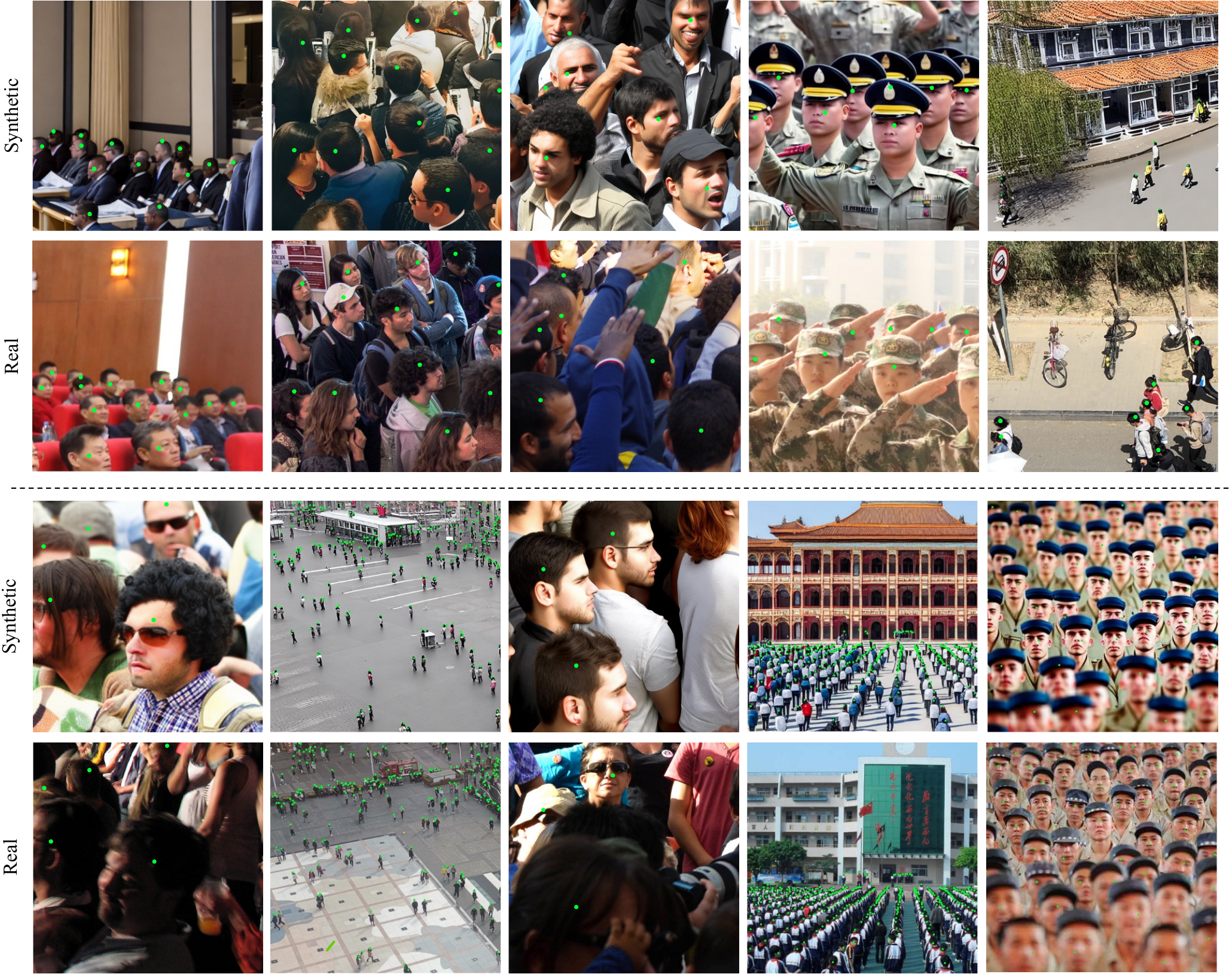}
    \caption{\textbf{Comparative Analysis of Real vs. Synthetic Crowd Images given Crowd Position Map}: Samples from the NWPU dataset, with \textcolor{green}{green} points indicating pre-determined crowd positions. The crowd distribution in the real image and corresponding synthetic image is the same.
    {Please Zoom in for details.} }
    \label{fig:visualization}
    \vspace{-10pt}
\end{figure*}

\subsection{Labeling-Free Data Augmentation}
Data augmentation with generative models hasn't been studied in the crowd counting research community for mainly two reasons: 1) For counting problem data augmentation, the synthetic data need to have a precise correspondence with their position annotations. Previously, there was no available conditional generation model that could achieve this. 2) Generative models in general are not good at generating high-frequency images with dense objects, while the datasets for most counting problems are dense. For example, they are good at generating images of one person with realistic details, but they are not good at generating hundreds of people in the same image.

In this work, we seek to produce high-quality training data for counting models. Since we do not have a training set large enough to train such a model from scratch, we rely on the pre-trained Stable Diffusion model, which we fine-tune for the crowd-counting task. During the training, the weights of Stable Diffusion are copied to ControlNet and are frozen, and we only train the copy of Stable Diffusion and the added zero convolution layers in ControlNet. However, this is not usually enough to produce high-quality images. To address this problem, we rely on conditioning, and investigate how to extend the ControlNet for crowd image generation conditioned on a head location map.\\
\textbf{Crowd-Counting Augmentation Pipeline.} 
For any given crowd dataset, our pipeline involves the following steps: 1) Train ControlNet on the crowd dataset with our proposed counting loss . The input to ControlNet is the density map generated from the location dot map, and the conditional text input can be either generated by BLIP \cite{li2022blip} or fixed to "a photo of a crowd of people" (See Figure \ref{fig:training-pipeline} for more details). 2) Use the trained ControlNet with our proposed guided sampling method to generate synthetic crowd images. 3) Use both the original dataset and the synthetic dataset to train the counting model of interest. 4) Evaluate the performance of the counting model on the validation set. \\
\noindent \textbf{Density Map as Input for ControlNet.} 
The straightforward method of feeding location dot maps into ControlNet for image generation falls short, particularly due to the sparseness of head localization dot maps (See Table \ref{fig:dotmap}). To address this, we apply a Gaussian kernel with a small bandwidth to the binary localization map \cite{adaptive_dm}, which transforms the dot map into a density map. Using the density map as ControlNet's input allows for a more accurate match between the synthetic crowd images and the control input. The richer spatially contextual information provided by the density map significantly improves ControlNet's ability to learn the correct locations to generate crowd images.

In particular, given the dot locations $D$, the density map is generated by:
\begin{equation}
y(x) = \sum_{i=1}^N \mathcal{N}(x | D_i, \beta\mathrm{I}),
\end{equation}
where $y(x)$ is the density value at the location $x$. $\mathcal{N}(x | \mu, \Sigma)$ is the Gaussian distribution with mean $\mu$ and convariance matrix $\Sigma$. In our experiments, we use a fixed value of $\beta=4$, which ensures that the map accurately represents the density of the crowd.

\noindent\textbf{Training Objectives.}
A common approach to train a diffusion model to meet a condition $C$, is to introduce the conditioning variable $C$ as an additional input to the denoising network $\epsilon_\theta$ and replace (\ref{eq:Lu}) by the conditional loss function 
\begin{align}
L_{c} 
& = \|\epsilon - \epsilon_\theta(x_t,t;C )\|^2  \nonumber.
\end{align}

This enables the neural network to learn to model the distribution $p(x|C)$. In Stable Diffusion, the condition is a text prompt specifying the content of the generated image, denoted as $C_{\text{text}}$. We also adopt the text prompt condition in our model to control the generated object of interest: $C_{\text{text}} = ``\text{A photo of a crowd of people}"$. To do data augmentation on other object counting problems, we can simply make $C_{\text{text}} = ``\text{A photo of \{object name\}}"$.

 We rely on density maps mentioned in the previous section as conditioning variable $C_{\text{dmap}}$ that serves as the control input for the ControlNet architecture. Our conditional loss function  that includes $C_{\text{dmap}}$ as the conditional variable is now as follows:
\begin{align}
L_{c} & = \|\epsilon - \epsilon_\theta(x_t,t;C_{\text{text}}, C_{\text{dmap}} )\|^2  
\end{align}

To improve the sensitivity of the diffusion model to this conditioning, we resort to a regularization loss that leverages an existing crowd counting model. Given image $x_t$ sampled at 
diffusion step $t$, we start by producing an estimate of the noise-free image ${\hat x}_{0|t}$, using (\ref{eq:xt}), i.e.
\begin{equation}
    \hat{x}_{0|t} = \frac{1}{\sqrt{\Bar{\alpha}_t}} (x_t - \sqrt{1 - \Bar{\alpha}_t}\epsilon).
    \label{eq:hatx}
\end{equation}
This image is then fed to a crowd counting model to produce an estimate  $\hat{y}_{\text{dmap}}( \hat{x}_{0|t})$ of the density map.
The loss function is then defined as 
\begin{equation}
L_{count} = \|y_{\text{gt}} - \hat{y}_{\text{dmap}}( \hat{x}_{0|t})\|^2  
\end{equation}
where $y_{\text{gt}}$ is the ground truth density map for $x_0$. 

However, recovering $x_0$ from $x_t$ produces blurred and noisy images when $t$ is large. While several techniques have been proposed to improve the quality of the reconstruction of $x_0$, such as mid-point estimation \cite{zhao2023diffswap}, we have not been able to successfully apply these techniques to crowd datasets, whose images tend to be much more complicated than those where they were originally tested (an image with only one cat/person, etc). To avoid the problem, we add the counting loss only when $t$ is small enough to generate clear crowd images. 
This leads to the final loss
\begin{equation}
    L = \begin{cases}
    L_{c} & 0 < t < t_{\text{threshold}}\\
    L_{c} + \lambda L_{count} & t_{\text{threshold}} < t < T
\end{cases},
\end{equation}
where $\lambda$ is a hyperparameter to guarantee that the two losses have roughly the same magnitude. $t_{\text{threshold}}$ is determined empirically, by visualizing how well $x_0$ is reconstructed from $x_t$. We set $t_{\text{threshold}} = 400$ in all our experiments. Our experiments show that this "weak guidance" loss turns out to be quite effective at enforcing the correspondence between the locations of the dot map and those of the heads in the synthesized image.

\begin{figure*}[t!]
    \centering
    \includegraphics[width=0.98\textwidth]{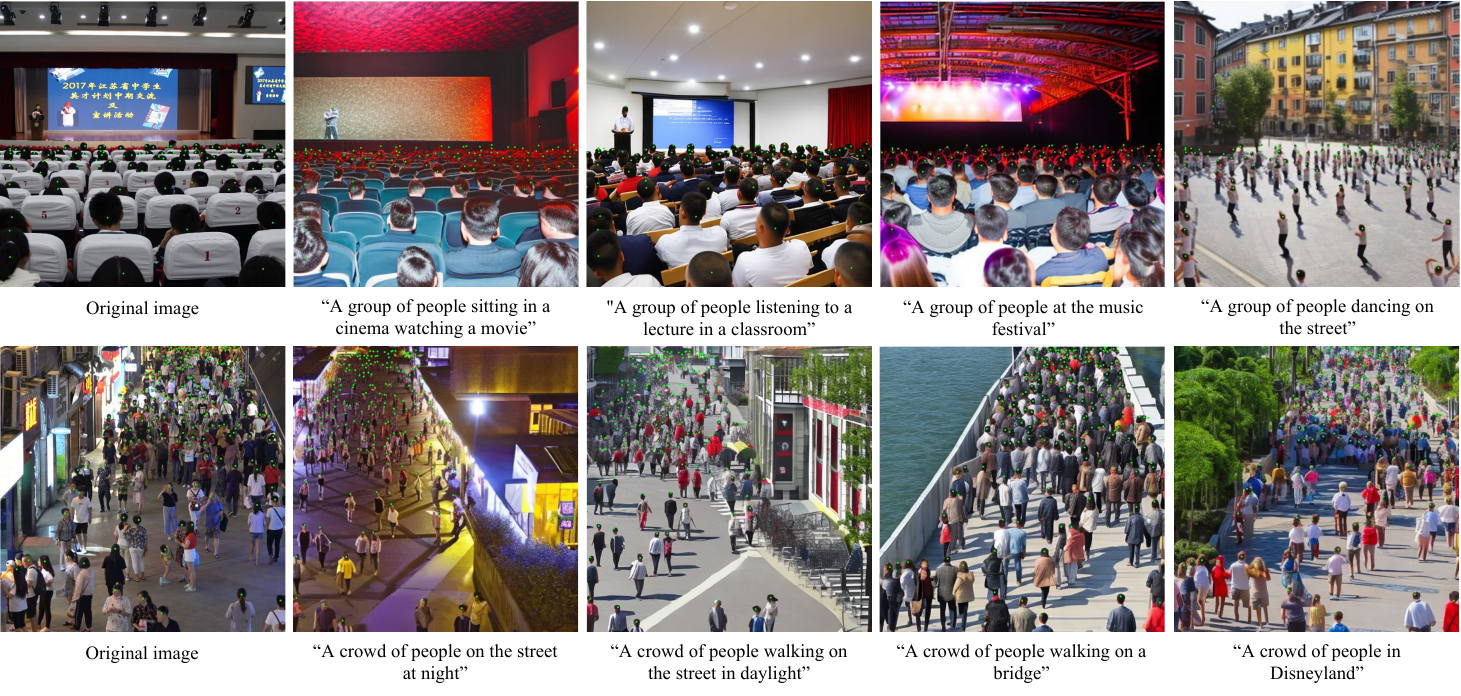}
    \caption{Generating crowd images given head location map and text prompts that control the background, with \textcolor{green}{green} points indicating pre-determined crowd positions. 
    Please Zoom in for details. }
    \label{fig:background}
    \vspace{-10pt}
\end{figure*}


\noindent\textbf{DDIM Sampling with Counting Guidance.} Inspired by the classifier-guided sampling approach \cite{dhariwal2021diffusion}, we propose a counting-guided sampling approach. This consists of updating the noise prediction $\epsilon_\theta(x_t,t,C_{\text{text}})$ at step $t$ with
\begin{dmath}
\tilde{\epsilon}(x_t) = \epsilon_\theta(x_t,t; C_{\text{text}}, C_{\text{dmap}}) - \sqrt{1 - \bar{\alpha_t}} \nabla_{x_t} \log p{(y_{\text{gt}}|x_t)},
\end{dmath}
where $\nabla_{x_t} \log p{(y_{\text{gt}}|x_t)}$ is the score function of the ground truth density map distribution. The gradient of the counting loss with respect to the input noisy image is given by
\begin{align}
    \nabla_{x_t} \log p{(y_{\text{gt}}|x_t)} &=  -\nabla_{x_t} \|y_{\text{gt}} - \hat{y}_{\text{dmap}}( \hat{x}_{0|t})\|^2.  
    \label{eq:score}
\end{align}

\begin{algorithm}
        \caption{Counting-Guided Diffusion Sampling}
        \begin{algorithmic}[1]
        \State Input: density map $y$, gradient scale $s$
        \State $x_T \leftarrow $ sample from $N(0,I)$
        \For{t from T to 1}
        \State compute $\Tilde{\epsilon}(x_t)$ with (\ref{eq:score}) - (\ref{eq:s})
        \State $x_{t-1} \leftarrow \sqrt{\bar{\alpha}_{t-1}(\frac{x_t - \sqrt{1 - \bar{\alpha}} \Tilde{\epsilon}}{\sqrt{\bar{\alpha_t}}})} + \sqrt{1 - \bar{\alpha}_{t-1}}\Tilde{\epsilon}$
        \EndFor
        \State \Return $x_0$
        \end{algorithmic}
        \label{algo}
\end{algorithm}
\vspace{-5pt}

In \cite{dhariwal2021diffusion}, classifier guidance is originally developed for classification tasks and applied to classifiers trained with noisy images. This relies on the robustness of classifiers to noise since class labels can frequently be inferred from global image information. However, counting models, which mostly analyze image details, are much more sensitive to noise. We observe that a counting model trained on clean images significantly outperforms one trained on noisy images in noisy image counting tasks. Hence, we use a counting model trained on clean images in all our experiments and adopt a dynamic guidance scale to alleviate its reduced accuracy on highly noisy images. This is implemented as
\begin{equation}
\tilde{\epsilon}(x_t) = \epsilon_\theta(x_t,t,C_{\text{text}}, C_{\text{dmap}}) - \alpha \sqrt{1 - \bar{\alpha_t}} \nabla_{x_t} \log p{(y_{\text{gt}}|x_t)},
\label{eq:guidance}
\end{equation}
where
\begin{equation}
\alpha = \frac{T-t}{T} \times s,
\label{eq:s}
\end{equation}
and $s$ is a constant that is set to  $s = 0.1$ in all our experiments.
In this way, counting scores of higher accuracy ($t$ small) provide more guidance to the sampling process. The proposed guidance sampling procedure is summarized by Algorithm~\ref{algo}.

\noindent\textbf{Text Prompt.} We also experiment with using BLIP ~\cite{li2022blip} to provide a text prompt for each image, instead of fixing the prompt $C_{text}$ to "a photo of a crowd of people". We find that BLIP will produce more detailed images compared to a fixed prompt. However, we suggest randomly deactivating the prompt condition during training if using BLIP, otherwise the ControlNet will entangle the prompt information with the crowd location information, resulting in inaccurate location correspondence during inference when text prompts are changed. In our experiments, we randomly deactivated the BLIP prompt with a ratio of 20\%.

\section{Experiments}
\label{sec:exps}

\subsection{Settings}

\textbf{Datasets.} Our augmented models are comprehensively trained and evaluated on three widely used crowd counting datasets separately: ShanghaiTech \cite{liu2018ano_pred}, which consists of 1198 images split into SHHA (482 images from web) and SHHB (716 images taken in Shanghai); UCF-QNRF \cite{idrees2018composition}, comprising 1535 high-resolution images that are collected from the web; and NWPU-Crowd \cite{gao2020nwpu}, which is currently the largest counting dataset containing 5,109 images and over 2.1 million annotated heads.


\noindent \textbf{Counting Models.} We selected counting models based on their performance on the NWPU-Crowd test set, which is currently the largest unpublished test dataset. We chose 3 open-sourced models that ranked the highest considering their MSE/MAE on this test set. The models we selected are: STEERER \cite{han2023steerer}, CUT \cite{CUT}, Generalized Loss (GL) \cite{Wan_2021_CVPR}.

\noindent \textbf{Implementation Details.} We used the vanilla ControlNet \cite{zhang2023adding} with $4 \times 64 \times 64$ latent space, and used the weight of Stable Diffusion (version 1.5) \cite{rombach2022highresolution} as the initial weight for training the ControlNet. We used the Adam \cite{kingma2017adam} optimizer with a learning rate initialized to $2\times10^{-5}$. During inference, we used DDIM sampler \cite{song2022denoising} with 50 sampling steps. We trained ControlNet on 2 GeForce RTX 3090 Ti, and used 1 NVIDIA RTX A6000 for sampling. For each image in the training set, we generated three $512 \times 512$ synthetic images. When training counting models with synthetic data, all the hyperparameters and settings are kept the same as reported in these counting papers.

\noindent \textbf{Metrics.} Followed by previous works \cite{BL}, the Mean Absolute Error (MAE) and Mean Squared Error (MSE) are used to evaluate the performance:
\begin{equation}
    MAE=\frac{1}{N}\sum_i\|\hat{y_i}-y_i\|, MSE=\sqrt{\frac{1}{N}\sum_i\|\hat{y_i}-y_i\|^2},
\end{equation}
where $N$ is the number of images. $\hat{y_i}$ and $y_i$ are predicted and ground-truth crowd counts. 

\subsection{Results}
\label{sec:results}
\textbf{Qualitative Results.} We perform a qualitative analysis of our model on the NWPU-Crowd dataset \cite{gao2020nwpu}, which is currently the largest crowd counting dataset available. As shown in Figure \ref{fig:visualization}, the location of the crowd in the generated images is the same as the location of the crowd in the real images. However, people's poses, expressions, and backgrounds are different, thus increasing the diversity of the training dataset. Moreover, we demonstrated in Figure \ref{fig:density} that our model is able to generate images with the correct crowd location at various density levels. In challenging high-density level cases, where all the people's heads are hard to discern even by human eyes, our model can still produce realistic images with crowd distribution similar to the original image.

\begin{table*}[h]
\centering
\caption{Comparison of different methods on SHHA, SHHB, UCF-QNRF, NWPU-Crowd datasets.}
    \resizebox{0.8\textwidth}{!}{
        \begin{tabular}{@{}llcccccccc@{}}
        \toprule
        \multirow{2}*{Method} & \multirow{2}*{Venue} & \multicolumn{2}{c}{SHHA} & \multicolumn{2}{c}{SHHB} & \multicolumn{2}{c}{UCF-QNRF} & \multicolumn{2}{c}{NWPU-Crowd}\\ \cline{3-10} 
        & & MAE  & MSE  & MAE  & MSE  & MAE & MSE & MAE & MSE  \\ \midrule
        CAN \cite{CAN} & CVPR19 & 62.3 & 100.0 & 7.8 & 12.2 & 107.0 & 183.0 & 106.3 & 386.5\\
        BL \cite{BL} & ICCV19 & 62.8 & 101.8 & 7.7 & 12.7 & 88.7 & 154.8 & 105.4 & 454.2\\
        DM-Count \cite{DM-Count} & NeurIPS20 & 59.7 & 95.7 & 7.4 & 11.8 & 85.6 & 148.3 & 88.4 & 388.6\\
        GL \cite{Wan_2021_CVPR} & CVPR21 & 69.8 & 142.4 & 7.3 & 11.7 & 84.3 & 147.5 & 79.3 & 346.1\\
        D2CNet \cite{D2CNet} & IEEE-TIP21 & 57.2 & 93.0 & 6.3 & 10.7 & 81.7 & 137.9 & 85.5 & 361.5\\
        P2PNet \cite{P2PNet} & ICCV21 & 52.7 & 85.1 & 6.3 & 9.9 & 85.3 & 154.5 & 72.6 & 331.6\\
        Chfl \cite{Chfl} & CVPR22 & 57.5 & 94.3 & 6.9 & 11.0 & 80.3 & 137.6 & 76.8 & 343.0\\
        CLTR \cite{liang2022endtoend} & ECCV22 & 56.9 & 95.2 & 6.5 & 10.6 & 85.8 & 141.3 & 74.3 & 333.8\\
        CHS-Net \cite{dai2023cross} & ICASSP23 & 59.2 & 97.8 & 7.1 & 12.1 & 83.4 & 144.9 & $-$ & $-$ \\
        DMCNet \cite{wang2023dynamic} & WACV23 & 58.5 & 84.5 & 8.6 & 13.7 & 96.5 & 164.0 & $-$ & $-$ \\  \midrule
        CUT \cite{CUT} & WACV21 & 55.7 & 86.3 & 7.9 & 12.9 & 84.4 & 156.5 & 69.3  & 304.0 \\
        \color{teal}{Ours (CUT)} & Ours & 53.9 & 85.2 & 7.1 & 11.8 & 82.0 & 145.5  & 68.4 & \textbf{259.2}\\ \midrule
        GL \cite{Wan_2021_CVPR} & CVPR21 & 61.3 & 95.4 & 7.3 & 11.7  & 84.3 & 147.5 & 79.3 & 346.1 \\
        \color{teal}{Ours (GL)} & Ours & 60.1 & 93.5 & 6.7 & 11.0  & 80.8 & 142.5 & 79.2 & 337.8\\ \midrule
        STEERER \cite{han2023steerer}& ICCV23 & 54.5 & 86.9 & \textbf{5.8} & \textbf{8.5} & 77.8 & 138.0 & 66.8 & 323.4 \\
        \color{teal}{Ours (STEERER)} & Ours & \textbf{52.3} & \textbf{82.3} & 
         5.9 & 8.9   & \textbf{77.2} & \textbf{130.9} &\textbf{64.7} & 310.5 \\ 
        \bottomrule
        \end{tabular}
    }
    \label{table:method_comparison}
\vspace{-10pt}
\end{table*}

\noindent \textbf{Quantitative Results on Counting Performance.} We compare the performance of the counting models when using only real data and when using both real and synthetic data. We demonstrate that training with synthetic data improves the counting models' performance on almost all the counting metrics (see the last 6 rows in Table \ref{table:method_comparison}). Also, our data-augmented models achieve state-of-the-art performance in all the counting datasets (see Table \ref{table:method_comparison}). Additionally, We notice that our data augmentation method is more effective in reducing the MSE score compared to the MAE score (e.g. the performance of the data augmented CUT on the NWPU-Crowd dataset). This implies that our model exhibits greater stability across all the test images, as evidenced by the absence of instances where the model produces an excessively large counting error that leads to a high MSE score. Moreover, we investigate the performance of our data-augmented models at various crowd density levels using the CUT model as an example. According to Table \ref{table:count_density}, our synthetic data improve the performance of the CUT model when the ground-truth people count is $n < 400$ or $n > 800$.

\begin{table}[h]
\centering
\caption{Comparison of ``CUT'' and ``Ours (CUT)'' (synthetic data augmented CUT) at different crowd density levels, where $n$ is the ground-truth crowd count.}
\resizebox{0.48\textwidth}{!}{
\begin{tabular}{@{}lccccccccc@{}}
\toprule
\multirow{2}*{}& \multicolumn{2}{c}{$n<$400} & \multicolumn{2}{c}{400 $\leq n <$800} & \multicolumn{2}{c}{ $n \geq$ 800} \\ \cline{2-7}
 & MAE & MSE & MAE & MSE & MAE & MSE \\
\midrule
CUT & 27.1 & 154.4 & \textbf{68.9} & \textbf{104.1} & 357.8 & 737.6 \\
Ours (CUT) & \textbf{22.5} & \textbf{50.9} & 77.4 & 119.0 & \textbf{355.8} & \textbf{727.3}\\
\bottomrule
\end{tabular}
}
\label{table:count_density}
\vspace{-10pt}
\end{table}

\begin{figure}[t]
\includegraphics[width=0.45\textwidth]{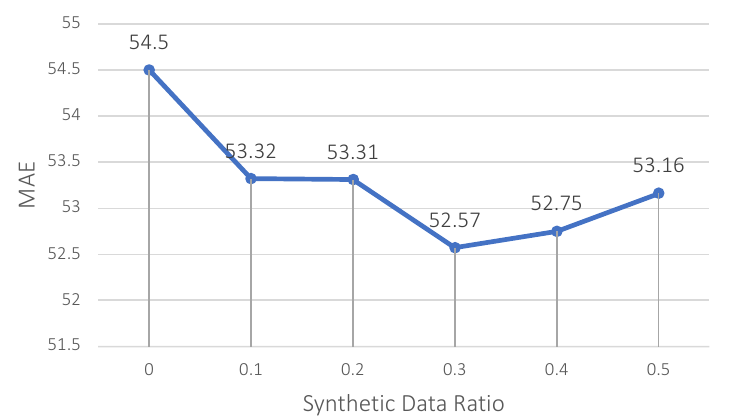}
\caption{Synthetic data ratio effect on MAE for the STEERER model on ShanghaiTech Part A.}
\label{fig:ratio}
\vspace{-10pt}
\end{figure}

\noindent\textbf{Ablation Study on Synthetic Data Ratio.} We investigate how the ratio of synthetic data affects the counting models' performance. Increasing the synthetic data ratio enhances the diversity of the training data, thereby making the models more generalizable to the test set. However, there is still a quality difference between the synthetic data and the real data, so adding too much synthetic data can compromise the overall quality of the training data. Therefore, it is crucial to find a balance point in the ratio of synthetic data to real data. In our experiments, we use STEERER on ShanghaiTech Part A dataset. We progressively increase the synthetic data sampling ratio from 0\% to 50\%. When the ratio is 50\%, it means that in each batch, there is an equal chance of sampling both real data and synthetic data. We empirically find that keeping the synthetic data ratio at around 30\% produces the best result on the test set (see Figure \ref{fig:ratio}).



\begin{figure}[t!]
\includegraphics[width=0.47\textwidth]{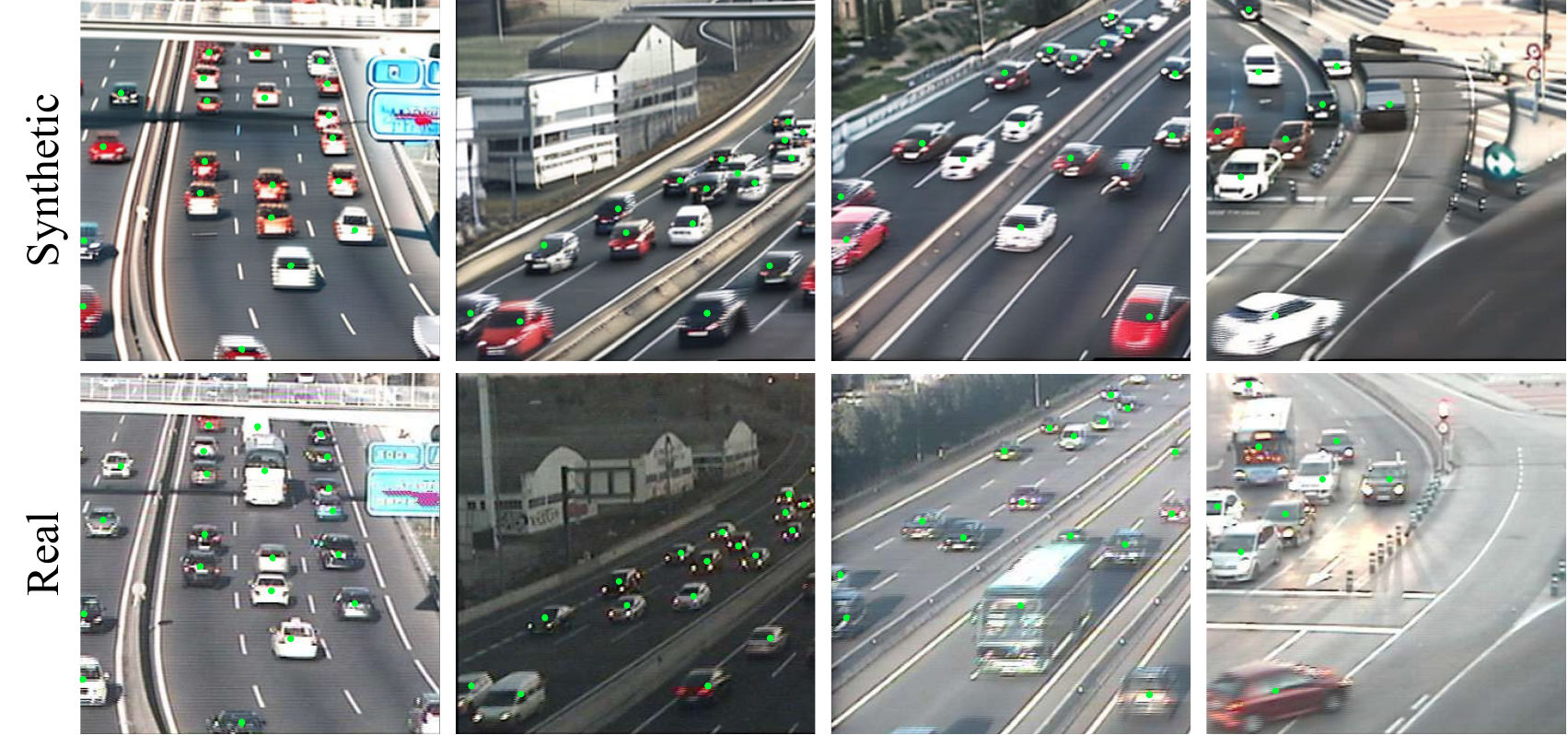}
\caption{Generating traffic congestion scenes with TRANCOS dataset, with \textcolor{green}{green} points indicating pre-determined car locations.}
\label{fig:transcos}
\vspace{-10pt}
\end{figure}

\begin{figure}[t]
    \centering
    \includegraphics[width=0.49\textwidth]{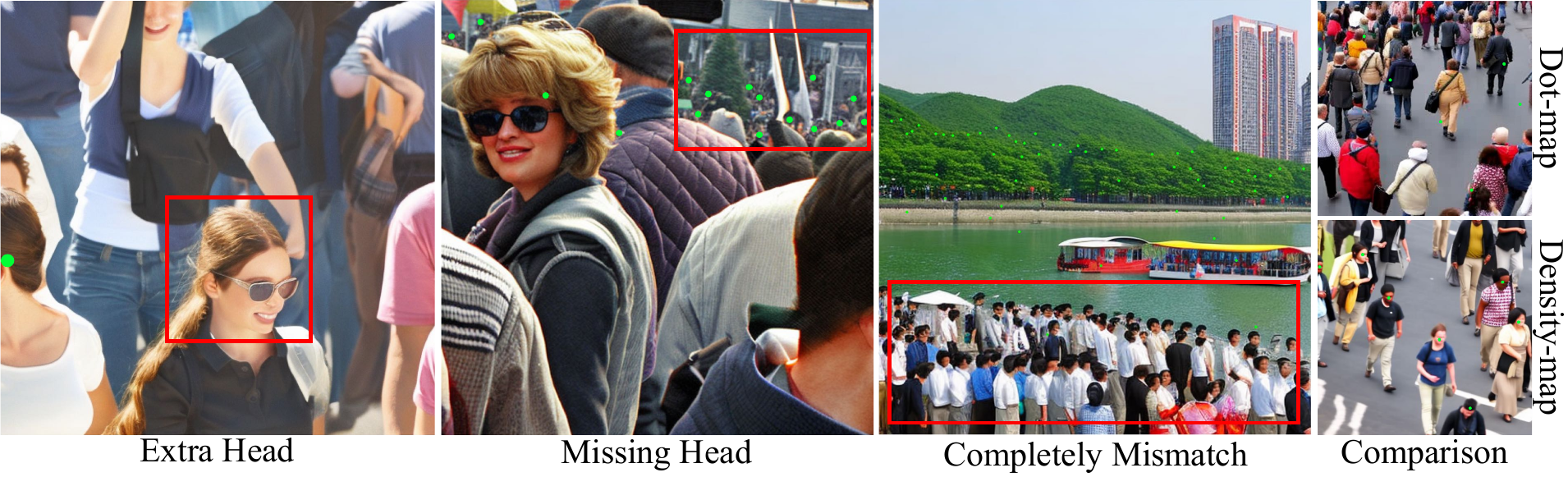}
    \caption{Undesirable results using dot maps as conditioning.}
    \label{fig:dotmap}
    \vspace{-10pt}
\end{figure}

\begin{table}[h]
\centering
\caption{Results on TRANCOS}
\resizebox{0.48\textwidth}{!}{
\begin{tabular}{@{}lcccc@{}}
\toprule
Methods & CSRNet \cite{li2018csrnet} & RSI-ResNet \cite{cheng2022rethinking} & STEERER & Ours (STEERER)\\
\midrule
MAE & 3.6 & 2.1 & 1.8 & \textbf{1.7}\\
MSE & - & 2.6 & 3.1 & \textbf{3.0}\\
\bottomrule
\end{tabular}
}
\label{table:trancos_table}
\vspace{-10pt}
\end{table}

\noindent \textbf{Vehicle Counting.} We demonstrate the versatility of our framework by conducting tests on various objects beyond just crowds, thus showcasing its applicability to a wide range of counting scenarios. We observe that crowd counting is one of the most challenging counting problems due to its exceptionally high density, so it's reasonable to assume that our framework works well on other less dense counting datasets as well. As part of our ablation study, we use the TRANCOS dataset \cite{TRANCOSdataset_IbPRIA2015}, which consists of surveillance photos capturing scenes of traffic congestion. We use STEERER's \cite{han2023steerer}
released checkpoints on the TRANCOS dataset for counting loss and counting guidance calculation. The result shows that the generated images correspond well with the binary car location map as shown in Figure \ref{fig:transcos}. The data-augmented counting result is shown in Table \ref{table:trancos_table}. There's little room for improvement, but we still improve both the MSE and MAE by 0.1.

\noindent\textbf{Background Controllable Crowd Generation.} Our framework is also flexible enough to control the backgrounds of the generated crowd images while ensuring the correctness of crowd location. This capability is achieved through the integration of BLIP \cite{li2022blip} for generating descriptive image captions as text prompts during training. Additionally, these text prompts are randomly deactivated at a rate of 0.2, thereby allowing ControlNet to learn a more diverse range of textual information and preventing it from overfitting to text prompts. During sampling, we simply need to change the input text prompt to "a crowd of people + {background description}". This enables us to augment the counting data with any desired background. Visualization results are included in Figure \ref{fig:background}.

\noindent\textbf{Dotmap v.s. Density Map.}
To assess our smooth conditioning approach, we analyze images generated using dot map conditioning, as shown in Figure~\ref{fig:dotmap}. Here, we observe a mismatch between the generated images and the dot maps, where some extra heads appear in unintended areas, and some heads are missing for annotation points. However, images conditioned using the proposed method show correspondence with the dot maps, as in Figure~\ref{fig:visualization}.

\noindent \textbf{Limitations.}
By modifying the loss function and adding guidance sampling, from a mathematical perspective, we are sacrificing the image quality with the accurate correspondence between dotmap and crowd position. In the future, more work can be done on improving generated image quality while enforcing this correspondence.

\section{Conclusion}
We introduce an innovative framework that augments the training data for object counting problems. Several efforts have been made to use the diffusion model for counting data augmentation tasks, including designing smooth conditional inputs, counting loss functions, and counting guided sampling methods. Our method significantly improves counting model performance and enhances generalization, as demonstrated by extensive experiments. We hope our work can inspire more people to explore how to use generative models to increase training data diversity on counting tasks. 
{
    \small
    \bibliographystyle{ieeenat_fullname}
    \bibliography{main}
}


\end{document}